\documentclass[sigconf]{aamas}

\usepackage{balance} 

\usepackage{amsmath, mathtools, geometry}
\usepackage{xcolor}
\usepackage{algorithm}
\usepackage{algpseudocode}
\usepackage{amsthm}
\theoremstyle{definition}

\usepackage{amsmath,bbm}       
\DeclareMathOperator*{\argmax}{arg\,max}
\usepackage{hyperref}
\usepackage{booktabs}
\usepackage{graphicx}
\usepackage{subcaption}
\usepackage{multirow}
\usepackage{float}
\usepackage{balance} 
\usepackage{enumitem}
\usepackage{caption}
\usepackage{placeins}
\usepackage{bbm}
\setlist[itemize]{topsep=2pt, itemsep=1pt, parsep=0pt, partopsep=0pt, leftmargin=*}

\doi{GWFE6009}

\makeatletter
\gdef\@copyrightpermission{
  \begin{minipage}{0.2\columnwidth}
   \href{https://creativecommons.org/licenses/by/4.0/}{\includegraphics[width=0.90\textwidth]{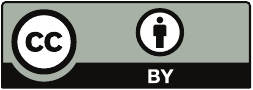}}
  \end{minipage}\hfill
  \begin{minipage}{0.8\columnwidth}
   \href{https://creativecommons.org/licenses/by/4.0/}{This work is licensed under a Creative Commons Attribution International 4.0 License.}
  \end{minipage}
  \vspace{5pt}
}
\makeatother

\setcopyright{ifaamas}
\acmConference[AAMAS '26]{Proc.\@ of the 25th International Conference
on Autonomous Agents and Multiagent Systems (AAMAS 2026)}{May 25 -- 29, 2026}
{Paphos, Cyprus}{C.~Amato, L.~Dennis, V.~Mascardi, J.~Thangarajah (eds.)}
\copyrightyear{2026}
\acmYear{2026}
\acmDOI{}
\acmPrice{}
\acmISBN{}

\acmSubmissionID{1291}

\title[AAMAS-2026 Formatting Instructions]{Interpretable Failure Analysis in Multi-Agent Reinforcement Learning Systems}

\author{Risal Shahriar Shefin}
\authornote{These authors contributed equally to this research, and their orders are interchangeable.}
\affiliation{
  \institution{Wake Forest University}
  \city{Winston-Salem, NC}
  \country{USA}}
\email{shefrs24@wfu.edu}

\author{Debashis Gupta}
\authornotemark[1]
\affiliation{
  \institution{Wake Forest University}
  \city{Winston-Salem, NC}
  \country{USA}}
\email{guptd23@wfu.edu}

\author{Thai Le}
\affiliation{
  \institution{Indiana University}
  \city{Bloomington, IN}
  \country{USA}}
\email{tle@iu.edu}

\author{Sarra Alqahtani}
\affiliation{
  \institution{Wake Forest University}
  \city{Winston-Salem, NC}
  \country{USA}}
\email{sarra-alqahtani@wfu.edu}

\begin{abstract}

Multi-Agent Reinforcement Learning (MARL) is increasingly deployed in safety-critical domains, yet methods for interpretable failure detection and attribution remain underdeveloped. We introduce a two-stage gradient-based framework that provides interpretable diagnostics for three critical failure analysis tasks: (1) detecting the true initial failure source (\textsc{Patient-0}); (2) validating why non-attacked agents may be flagged first due to domino effects; and (3) tracing how failures propagate through learned coordination pathways. Stage~1 performs interpretable per-agent failure detection via Taylor-remainder analysis of policy-gradient costs, declaring an initial \textsc{Patient-0} candidate at the first threshold crossing. Stage~2 provides validation through geometric analysis of critic derivatives-first-order sensitivity and directional second-order curvature aggregated over causal windows to construct interpretable contagion graphs. This approach explains ``downstream-first'' detection anomalies by revealing pathways that amplify upstream deviations. Evaluated across 500 episodes in Simple Spread (3 and 5 agents) and 100 episodes in StarCraft~II using MADDPG and HATRPO, our method achieves \mbox{88.2--99.4\%} \textsc{Patient-0} detection accuracy while providing interpretable geometric evidence for detection decisions. By moving beyond black-box detection to interpretable gradient-level forensics, this framework offers practical tools for diagnosing cascading failures in safety-critical MARL systems.

\end{abstract}


\newcommand{\BibTeX}{\rm B\kern-.05em{\sc i\kern-.025em b}\kern-.08em\TeX}

\begin{document}


\pagestyle{fancy}
\fancyhead{}


\maketitle 
\section{Introduction}
Multi-Agent Reinforcement Learning (MARL) has emerged as a pivotal framework for coordinating autonomous systems across safety-critical domains including robotic fleets, transportation systems, communication networks, and power grids~\cite{petrenko2021multi,cps,Minrui,Panfili, Peng,Noureddine2017MultiagentDR}. In these settings, where multiple learned policies interact, a single local deviation can ripple through the entire team causing other agents to fail~\cite{MARLPrima}. 

However, understanding \emph{how} failures propagate in MARL systems presents unique challenges. In tightly coupled multi-agent systems, the first agent to show detectable instability may not be the true failure source; much like a domino effect where the first domino pushed (true source) might fall slowly, while a downstream domino (affected agent) falls dramatically due to amplification. This phenomenon creates fundamental diagnostic challenges: an agent might be flagged as failing first simply because it's more sensitive to perturbations, while the true root cause remains hidden upstream. In such circumstances, practitioners need explanations that reveal three critical aspects: \emph{who} truly deviated first, \emph{why} certain agents were affected before others, and \emph{how} the deviation propagated through the system. We posit that answers to these questions are paramount and that \emph{safety in MARL is inseparable from explainability and transparency}: only by making agents' behavior and their interdependencies legible can practitioners debug, test, and ultimately trust these systems.

Despite these needs, explainable MARL (xMARL) remains underexplored. Shapley-value attribution~\cite{shap, shap2} has been used to quantify each agent's contribution to team reward in cooperative scenarios. Although such attribution offers insights into credit assignment, it does \textit{not} reveal inter-agent dynamics or cascading failures. Complementary work~\cite{temporal} provides policy-level contrastive explanations by checking user-specified temporal task queries against abstracted policies via probabilistic model checking. While valuable, these approaches are not designed to analyze failure dynamics in team settings, such as how failures unfold, which agent was the inciting point of failure, and why.

In contrast, single-agent Explainable RL (XRL) has recently shifted toward safety debugging and testing, using explanations to surface vulnerabilities and guide falsification (e.g., \cite{xsrl,falsification}). For instance, xSRL~\cite{xsrl} combines local post-hoc explanations with global abstract policy graphs to identify and patch safety violations. AEGIS-RL~\cite{falsification} builds on xSRL by developing a verification approach that integrates formal model checking with risk- and uncertainty-guided falsification. However, repurposing these single-agent methods for MARL is not straightforward due to the heightened complexity of multi-agent systems. The explainability challenge in MARL surpasses that of single-agent RL, primarily due to the inherent coupling and dependencies among agents. While single-agent explanations can focus on an individual policy, MARL requires untangling a web of interdependencies, compounded by emergent collective behaviors and environmental non-stationarity. These dynamic feedback loops obscure cause-effect relationships, making it difficult to distinguish between agent-level and system-level failures. 

Moreover, conventional failure analysis in MARL often tracks global metrics such as team reward or derived statistics over time~\cite{performance1, performance2}. Such global performance metrics can obscure policy instability in individual agents~\cite{Abdallah2009WhyGP,performance3} as we show in our experiments. Moreover, they neither identify the first agent to enter a non-robust state nor reveal how deviations propagate, all of which are needed to pinpoint weak points and pathways for proactive mitigation.

Therefore, this paper formalizes the interpretable failure analysis problem in MARL around three core questions; \textbf{Q1.} Who is the true \textsc{Patient-0}: the first agent to enter a non-robust state?, \textbf{Q2.} Why might a non-attacked agent be flagged first? How do we distinguish true sources from amplified downstream effects?, and \textbf{Q3.} How does instability propagate across agents over time through the system's learned coordination pathways? We propose to address these questions through a two-stage, gradient-based forensic method. Stage~1 performs per-agent local detection by probing the Taylor-error of a policy-gradient cost; the earliest threshold crossing declares an initial \textsc{Patient-0} candidate (Q1). Stage~2 validates or corrects this candidate by tracing accelerating upstream influence using critic derivatives-first- and second-order action sensitivities aggregated over causal windows (Q2). This approach naturally yields constructions of directed contagion graphs that summarize propagation pathways, influence strength, and timing (Q3).

\noindent\textbf{Contributions.} We make four main contributions:
\begin{enumerate}
  \item \textbf{Two-stage \textsc{Patient-0} detection.} A gradient-based method that combines local policy instability detection with upstream traceback validation.
  \item \textbf{Geometric explanation of detection anomalies.} Directional second-order analysis distinguishes accelerating vs. damping influence, explaining why downstream agents might be flagged before true sources.
  \item \textbf{Interpretable contagion graphs.} Directed graphs that quantitatively capture influence strength, amplification frequency, and propagation timing.
  \item \textbf{Comprehensive evaluation.} Across two environments and two algorithms, we report \textsc{Patient-0} identification accuracy of $88.2\%-99.4\%$ with intervention studies validating that perturbing high-influence, accelerating states induces significantly longer instability.
\end{enumerate}

By moving beyond team reward to gradient-level influence analysis, our approach provides actionable insight into systemic vulnerabilities, revealing not just which agent failed first, but \emph{why} and \emph{through which pathways} failures spread, enabling more reliable MARL testing and deployment in safety-critical domains.
\section{Related Work}
\noindent \textbf{Explainability in MARL (xMARL)}. Research in explainable MARL is limited and primarily splits into (i) credit assignment~\cite{shap,shap2} and (ii) specification- or query-driven explanations~\cite{temporal}. Shapley-based methods~\cite{shap,shap2} quantify each agent's contribution to the team reward, offering insights into fairness and cooperation but not into failure dynamics or inter-agent cascades . Recent work continues to advance credit assignment, such as Multi-level Advantage Credit Assignment (MACA)~\cite{MACA}, which performs counterfactual reasoning across different levels of agent coordination but remains focused on optimizing training rather than diagnosing post-hoc failures . Complementarily, policy-level contrastive explanations ~\cite{temporal} verify temporal task queries (e.g., encoded in PCTL*) via probabilistic model checking on an abstracted policy . While both lines illuminate aspects of MARL behavior, neither targets post-hoc failure analysis under compromised agents: they do not identify the entry point of a failure nor reconstruct how influence propagates across agents over time. Our method addresses this gap by detecting the first non-robust agent and mapping directed, time-resolved propagation paths using centralized-critic sensitivities.

\noindent \textbf{Single-agent XRL and Policy Summaries.} A rich body of XRL provides explanations for a single agent, including policy summarization via abstract policy graphs~\cite{summarize, caps_aamas22}, contrastive explaination methods~\cite{contrastive}, and frameworks that integrate local and global summaries~\cite{xsrl,falsification}. Safety-focused XRL (e.g., xSRL~\cite{xsrl}) ties explanations to debugging and patching unsafe behavior but remains single-agent in scope. Directly porting these methods to MARL misses the core challenge: the coupling and non-stationarity among agents obscure cause-effect relations across the team. Our work departs by explicitly modeling cross-agent influence during failure episodes and delivering a cascade map that explains how a local deviation spreads at run time.


\noindent \textbf{Attacks and Robustness.} The vulnerability of cooperative MARL to adversarial pressure is well-studied, including attacks that degrade team performance~\cite{MARLPrima, ijcai23_anomaly, ami_attack,Lischke}. These works establish vulnerabilities and propose defenses, but generally focus on performance degradation or anomalous behavior detection without explaining the failure's origin and propagation pathway. Our contribution is orthogonal and complementary: a post-hoc explainer that identifies \textsc{Patient-0} and attributes influence across agents.

\section{Approach}

We propose a two-stage, interpretable gradient-based method for MARL failure forensics. The key insight is that a failure typically begins with one agent and then propagates via learned coordination. Stage~1 performs per-agent local detection by probing policy-gradient Taylor error and declares \textsc{Patient-0} as the first threshold crossing. Stage~2 validates or corrects it by tracing accelerating upstream influence using critic derivatives-first-order sensitivity and directional curvature-aggregated over a short causal window. These scores yield a directed contagion graph that summarizes who influenced whom, the strength and frequency of amplification, and how quickly deviations spread. We assume a single failure entry per episode, with subsequent failures arising from cascades through cooperative coupling.\footnote{Relaxing the single–entry assumption is left to future work.}

\subsection{Two-Stage \textsc{Patient-0} Detection and Validation}
\label{subsec:twostage-p0}
In tightly coupled multi-agent systems, the first agent to show detectable instability may not be the true failure source. Consider a domino effect: the first domino pushed (true source) might fall slowly, while a downstream domino (affected agent) falls dramatically due to amplification. Stage 1 identifies which domino fell first; Stage 2 traces back to find who pushed it.


\vspace{2pt}
\noindent\textbf{Stage 1: Local Policy Instability via Taylor Remainder.}
We monitor each agent's policy stability by measuring sensitivity to small observation perturbations. The key intuition: when policies enter fragile states, small changes in observations cause disproportionately large changes in policy update directions.

For agent \(i\) with policy \(\pi_i(a\mid o_i)\), we define the action commitment cost:
\begin{equation}
\begin{aligned}
J_i(o_i,\tau_i) &= -\sum_{a} \tau_i(a)\,\log \pi_i(a\mid o_i),\\
\tau_i(a) &\equiv \pi_i^*(a\mid o_i^t)
           = \mathbbm{1}_{a = \argmax_{a'}\pi(a'\rvert o_i^t)}.
\label{eq:pgcost}
\end{aligned}
\end{equation}
By fixing \(\tau_i\) at the current greedy action, we transform policy evaluation into a deterministic function of observations. Gradients and Hessians are taken w.r.t.\ observations with \(\tau_i\) held fixed:
\[
\nabla_{o_i} J_i(o_i^t,\tau_i),
\qquad
\nabla_{o_i}^2 J_i(o_i^t,\tau_i).
\]

For a small perturbation \(\eta_i\), Taylor's theorem gives:
\[
J_i(o_i^t+\eta_i,\tau_i)
= J_i(o_i^t,\tau_i) + \nabla_{o_i}J_i(o_i^t,\tau_i)^\top \eta_i
+\tfrac{1}{2}\eta_i^\top \nabla_{o_i}^2 J_i(\tilde o_i,\tau_i)\,\eta_i,
\]
for some \(\tilde o_i\in[o_i^t,o_i^t+\eta_i]\). We define the Taylor remainder statistic:
\begin{equation}   
\mathcal L_i^t(\eta_i)
\;\triangleq\;
J_i(o_i^t+\eta_i,\tau_i)
-\Big(J_i(o_i^t,\tau_i)+\nabla_{o_i}J_i(o_i^t,\tau_i)^\top \eta_i\Big),
\label{eq:taylor-error}
\end{equation}
so that for small \(\eta_i\), \(\mathcal L_i^t(\eta_i)\approx \tfrac{1}{2}\eta_i^\top \nabla_{o_i}^2 J_i(o_i^t,\tau_i)\eta_i\). This provides a curvature probe in observation space without forming the full Hessian, similar to the Second-Order Identification of Non-Robust Directions test from single-agent RL~\cite{soinrd}. Large deviations indicate the policy is in a non-robust, high-curvature region where small observation changes could cause large policy shifts.


This local per-agent policy curvature test is effective in MARL because during decentralized execution, each agent $i$ executes its policy based solely on local observations $o_i$, which implicitly encode teammates' behavior through environment dynamics. A deviation by any agent perturbs others' observation manifolds, altering the local geometry of $J_i(o_i)$. Thus, monitoring the directional curvature of $J_i$ with respect to $o_i$ serves as a policy-agnostic, per-agent detector of entry into non-robust state without requiring explicit inter-agent communication or access to peers' policies.

Then, from fault-free rollouts, we establish baseline stability profiles for each agent. During failure episodes, we flag agents when their Taylor-error significantly exceeds their normal operating range. The earliest-detected agent becomes our initial \textsc{Patient-0} candidate:
\begin{equation}
\label{eq:p0-stage1}
\hat p_{\text{S1}} \;=\; \arg\min_{i} T_i, \quad \hat T_{\text{attack}} \;=\; T_{\hat p_{\text{S1}}}.
\end{equation}
, where $T_i$ is the time when agent $i$ is flagged.

\vspace{6pt}
\noindent\textbf{Stage 2: Cascade Source Identification via Directional Critic Curvature.}
Stage 1 can be misled by downstream-first cascades, where a non-source agent crosses the instability threshold before the true origin. This happens when small upstream deviations are amplified at sensitive teammates, when observation coupling causes agents to indirectly “see” each other through the environment dynamics, or when the system passes through critical coordination states that magnify otherwise minor perturbations. To disambiguate these cases, we perform a directional second-order analysis on the critic: first-order sensitivities identify who most strongly affects whom, while the directional curvature of the critic isolates amplifying (vs. damping) states, allowing us to traceback the highest-leverage upstream source and validate or correct the Stage 1 \textsc{Patient-0}.

In MARL algorithms with action-value critic (e.g., MADDPG ~\cite{lowe2017multi}), we leverage the trained per-agent critic $Q_i(s,a_1,\dots,a_n)$ that conditions on the global state (or joint observations) and the joint action; this object encodes inter-agent dependencies learned during training. In algorithms that only learn a state-value function (e.g., HATRPO~\cite{hatrpo}), we fit a light post-hoc probe critic $\tilde Q_i(s,a_1,\dots,a_n)$ on frozen rollouts for attribution only (not for control), by regressing to TD or Monte-Carlo targets.\footnote{Either critic is required only to be differentiable with respect to each action argument so that we can take first- and second-order derivatives. For discrete actions we use a standard continuous relaxation (e.g., softmax with temperature or Gumbel–Softmax during probing); for continuous actions we directly backpropagate.} This yields well-defined partials $\partial Q_i/\partial a_j$ and $\partial^2 Q_i/\partial a_j^2$ that quantify how agent $j$’s action locally perturbs agent $i$’s value and, by extension, its policy’s stability signals. 
We define the critic cost function as \(L_{Q_i} = -Q_i\), thereby treating higher values as worse and lower values as better, which preserves the standard characteristics of a cost function.

For an ordered pair $(j\!\rightarrow\! i)$ at $(s_t,a_t)$, let
\[
\mathbf g_{ij}^t \;=\; \frac{\partial L_{Q_i}(s_t,a_t)}{\partial a_j},
\qquad
\mathbf H_{ij}^t \;=\; \frac{\partial^2 L_{Q_i}(s_t,a_t)}{\partial a_j^2\,}.
\]
We use the first-order influence magnitude $G_{ij}^t=\|\mathbf g_{ij}^t\|_2$ and the \emph{directional} second-order term
\begin{equation}
\label{eq:dir-second}
D_{ij}^t \;=\; \mathbf g_{ij}^{t\top}\,\mathbf H_{ij}^t\,\mathbf g_{ij}^t ,
\end{equation}
which evaluates curvature along the locally most sensitive direction. Under a second-order Taylor expansion of $\mathcal L_{Q_i}$ with respect to $a_j$,
\[
\Delta \mathcal L_{Q_i} \;\approx\; \mathbf g_{ij}^{t\top}\delta a_j \;+\; \tfrac{1}{2}\,\delta a_j^\top \mathbf H_{ij}^t \delta a_j .
\]
When $G_{ij}^t$ is large and $D_{ij}^t>0$, small deviations in $a_j$ are \emph{amplified} at $i$ (accelerating influence); when $D_{ij}^t<0$, they are \emph{damped} (saturating influence). This explains how a teammate $i$ can exhibit an earlier curvature spike than the true source agent (true \textsc{Patient-0} $j$.


To validate $\hat p_{\text{S1}}$ (Eq.\ref{eq:p0-stage1}) in Stage-1, we trace potential harmful upstream over a short window ending at the detection time, aggregating only the accelerating portions:
\begin{align}
\textstyle I_{j\to i}
&= \sum_{t=t_0}^{t_1} \omega(t_1{-}t)\, \mathbf 1\!\{D_{ij}^t>0\}\, |\mathcal L_j^t(\eta_j) - E\{\mathcal L_j^t\}| ,
\end{align}
where $[t_0,t_1]$ is a causal window (e.g., $t_1=T_i$ and $t_0=\max\{0,T_j\}$), $\omega(\cdot)$ is a recency weight, $I_{j\to i}$ is cumulative (accelerating) deviation impact, and $E\{\mathcal L_j^t\}$ denotes the expected baseline of $\mathcal L_j^t$ obtained from the previously computed stability profiles. We then iteratively move from the detected agent to its strongest accelerating upstream source until no stronger source is found (or a cycle would form).

\begin{algorithm}[t]
\caption{Traceback Validation of \textsc{Patient-0} with Directional Second Order}
\label{alg:traceback-validate}
\begin{algorithmic}[1]
\Require Stage-1 $\hat p_{\text{S1}}$ at time $t^\star$; window $K$; decay $\omega$
\State $\text{curr}\gets \hat p_{\text{S1}}$,\quad $\text{chain}\gets[\text{curr}]$,\quad $\mathcal W \gets \{t^\star{-}K,\ldots,t^\star\}$
\While{true}
  \ForAll{$j\neq \text{curr}$}
    \State $I_{j\to \text{curr}} \gets \sum_{t\in\mathcal W}\omega(t^\star{-}t)\,\mathbf 1\{D_{\text{curr},j}^t>0\}\, |\mathcal L_j^t(\eta_j) - E\{\mathcal L_j^t\}|$
  \EndFor
  \State $j^\star \gets \arg\max_j I_{j\to \text{curr}}$
  \If{$I_{j^\star\to \text{curr}}$ is negligible \textbf{or} $j^\star\in\text{chain}$}
     \State \textbf{break}
  \EndIf
  \State append $j^\star$ to $\text{chain}$;\quad $\text{curr}\gets j^\star$
\EndWhile
\State \Return $\text{chain}$ (upstream sources ending at $\hat p_{\text{S1}}$)
\end{algorithmic}
\end{algorithm}

\noindent The traceback (Algorithm~\ref{alg:traceback-validate}) hence validates Stage-1 by (i) revealing whether the Stage-1 flag lies on a high-influence, accelerating edge from an upstream agent, and (ii) identifying that upstream agent.

\vspace{0.5cm}
\noindent\textbf{Remarks.}
Stage-1’s Taylor-error $\mathcal{L}_i^t$ (Eq.~\eqref{eq:taylor-error}) probes directional curvature of the policy-gradient norm $J_i(o_i)$ (Eq.~\eqref{eq:pgcost}). Stage-2 traces cross-agent influence using critic derivatives $\mathbf g_{ij}^t=\partial Q_i/\partial a_j$ and $\mathbf H_{ij}^t=\partial^2 Q_i/\partial a_j^2$ as practical proxies for environmental coupling (under a mild critic–influence alignment on the analysis window). The directional curvature $D_{ij}^t=\mathbf g_{ij}^{t\top}\mathbf H_{ij}^t\mathbf g_{ij}^t$ (Eq.~\eqref{eq:dir-second}) decomposes as $D_{ij}^t=\|\mathbf g_{ij}^t\|_2^2\,\kappa_{ij}^t$ with $\kappa_{ij}^t$ the Rayleigh quotient. When $D_{ij}^t>0$ (accelerating states) and $G_{ij}^t=\|\mathbf g_{ij}^t\|_2$ is large, small perturbations $\delta a_j$ can disproportionately amplify downstream agents’ Taylor-errors, explaining ``downstream-first'' flags. Accordingly, our traceback aggregates only accelerating edges ($D_{ij}^t>0$) to validate or correct the Stage-1 candidate $\hat p_{\mathrm{S1}}$ (Eq.~\eqref{eq:p0-stage1}). Appendix~A provides a short lemma and proposition formalizing these links.

\subsection{Cross-Agent Action Influence for Cascade Attribution}
\label{subsec:influence}

The influence magnitude $G_{ij}^t$, and directional derivative $D_{ij}^t$ naturally yield a weighted, directed graph where edges $j\!\rightarrow\! i$ represent influence pathways. Edge weights combine $G_{ij}$ (first-order sensitivity) and the frequency of accelerating states ($D_{ij}^t > 0$). By thresholding on cumulative influence $I_{j\to i}$, we extract the salient contagion subgraph that explains failure propagation.

Let $t_i$ denote agent $i$'s detection time (first Taylor-error exceedance). For each subsequently failing agent $k$ and candidate predecessor $j$ with $t_j\le t_k$, we define the attribution window:
\[
\mathcal W_{j\to k} \;=\; \{\,t_j,\;t_j{+}1,\dots,t_k\,\}.
\]
This focuses analysis on the interval between upstream detection and downstream failure.

For each pair $(j\!\to\!k)$, we compute three interpretable summaries over $\mathcal W_{j\to k}$:
\begin{align}
 IS_{j\to k}
&\;=\;
\frac{1}{|\mathcal W_{j\to k}|}\sum_{t\in\mathcal W_{j\to k}}
G_{k j}^t,
\\[4pt]
CR_{j\to k}
&\;=\;
\frac{100}{|\mathcal W_{j\to k}|}
\sum_{t\in\mathcal W_{j\to k}}
\mathbf 1\!\left\{\,
G_{k j}^t \ge \theta_G
\ \wedge\
\tilde D_{k j}^t > 0
\right\},
\\[4pt]
\ t_{j\to k}
&\;=\; [t_k , t_j].
\end{align}
Here $\bar IS_{j\to k}$ is the influence score (magnitude), $CR_{j\to k}$ is the percentage of critical leverage steps (high influence with positive curvature), and $t_{j\to k}$ is the detection time period. Threshold $\theta_G$ is set to a robust scale (e.g., episode median).

We rank candidate predecessors using a score that combines our existing signals:
\[
S_{j\to k}
\;=\;
\sum_{t\in\mathcal W_{j\to k}}
\omega(t_k{-}t)\;
\max(\tilde D_{k j}^t,0)\;
G_{k j}^t,
\; \omega(\cdot)\in(0,1]\ \text{monotone}.
\]
We retain edge $j\!\to\!k$ if $S_{j\to k}$ exceeds fraction $\tau$ of $k$'s top score (e.g., $\tau\!=\!0.5$ for salient secondary parents).

We then construct a directed contagion graph over agents in detection order. Each node shows agent ID and policy instability occupancy $IO_i$; the number of timesteps the policy remained unstable due to incoming influences, measured from first to last Taylor-error exceedance. Each retained edge $j\!\to\!k$ reports $IS_{j\to k}\ ;\ \ CR_{j\to k}\% \ ;\\ t_{j\to k}\ \text{steps}$.

This contagion graph integrates all analytical components in a single view: direction (who influenced whom, via edge orientation), strength (total influence magnitude, \(IS\)), geometry (amplification frequency, \(CR\)), timing (the propagation period, \(t\)), and instability occupancy (\(IO\), the timesteps during which an agent’s policy was unstable under inbound influence). This integrated summary reveals both the severity of impact on individual agents and the mechanistic routes through which failures spread and amplify throughout the system.

\section{Experiments}

We evaluate our two-stage failure-analysis framework across multiple environments and algorithms to address three questions: \textbf{Q1:}~How accurately does the method identify the true \textsc{Patient-0}? \textbf{Q2:}~Why are non-attacked agents sometimes flagged first, and can traceback correct them? \textbf{Q3:}~Do the influence scores provide actionable, episode-level insight into how failures propagate?. All the codes relevant to the experiments are available online\footnote{\url{https://github.com/risal-shefin/marl-failure-analysis}}.

\subsection{Setup}

\noindent\textit{Environments and algorithms.}
We use Simple Spread (cooperative navigation) with $n\!\in\!\{3,5\}$ agents and SMAC (Starcraft II). In Simple Spread, $n$ agents must navigate to $L$ landmarks through coordinated movement. Agents observe relative positions of other agents and landmarks, and receive shared rewards to minimize distance to landmarks. SMAC is a decentralized micromanagement benchmark where 3 Marine agents must defeat 3 Zealot opponents through coordinated focus fire and positioning. Each agent receives partial observations and controls one Marine unit, requiring emergent coordination strategies to overcome the enemy's statistical advantages. We tested two MARL algorithms; MADDPG~\cite{lowe2017multi} and HATRPO~\cite{hatrpo}.  
\begin{table*}[t]
\centering
\caption{\textsc{Patient-0} identification. Columns (left to right): \textbf{Stage-1} accuracy; \textbf{Correction Rate} (computed only over downstream-first cases); and \textbf{Combined} accuracy after Stage-2 correction.}
\label{tab:p0-results}
\resizebox{0.53\linewidth}{!}{
\setlength{\tabcolsep}{6pt}
\begin{tabular}{l l c c c}
\toprule
\textbf{Setting} & \textbf{Algorithm} & \textbf{Stage-1} & \textbf{Correction Rate} & \textbf{Combined} \\
\midrule
\multirow{2}{*}{SimpleSpread-3} & MADDPG & 95.7\% & 66.9\% & 98.6\% \\
                                & HATRPO & 99.1\% & 66.7\% & 99.4\% \\
\cmidrule(lr){1-5}
\multirow{2}{*}{SimpleSpread-5} & MADDPG & 88.1\% & 40.1\% & 92.8\% \\
                                & HATRPO & 98.9\% & 48.6\% & 99.2\% \\
\cmidrule(lr){1-5}
\multirow{2}{*}{SMAC (3s\_v\_3z)} & MADDPG & 84.0\% & 70.8\% & 88.2\% \\
                                  & HATRPO & 94.8\% & 67.7\% & 98.3\% \\
\bottomrule
\end{tabular}
}
\end{table*}

\begin{table*}[t]
\centering
\caption{Influence metric accuracy: Critical state interventions cause significantly longer instability than robust state interventions across the paired scenarios. Our instability occupancy (IO) outperforms traditional performance metrics in revealing vulnerabilities.}
\label{tab:influence-validation}
\resizebox{0.53\linewidth}{!}{
\setlength{\tabcolsep}{6pt}
\begin{tabular}{l l c c c}
\toprule
\textbf{Setting} & \textbf{Algorithm} & \textbf{IO} & \textbf{AUC-Q} & \textbf{AUC-Reward} \\
\midrule
\multirow{2}{*}{Simple Spread (3 agents)} & MADDPG & 77.7\% & 51.2\% & 48.4\% \\
                                          & HATRPO & 74.0\% & 59.6\% & 62.3\% \\
\cmidrule(lr){1-5}
\multirow{2}{*}{Simple Spread (5 agents)} & MADDPG & 71.3\% & 49.9\% & 51.2\% \\
                                          & HATRPO & 82.4\% & 72.1\% & 56.6\% \\
\cmidrule(lr){1-5}
\multirow{2}{*}{SMAC (3s\_vs\_3z)} & MADDPG & 54.5\% & 53.2\% & 47.7\% \\
                                           & HATRPO & 59.7\% & 45.7\% & 52.7\% \\
\bottomrule
\end{tabular}
}
\end{table*}

\noindent\textit{Failure Simulation/ Intervention Protocol.}
We simulate failures using the worst-action attack from adversarial MARL~\cite{ami_attack,MARLPrima}. We then created an intervention protocol to test our proposed work. For each ordered pair $(j\!\to\!i)$, we intervene at two classes of timesteps computed from the trained critic(s): \emph{Critical} states where $G_{ij}$ is high and $D_{ij}\!>\!0$, and \emph{Robust} states where $G_{ij}$ is low and $D_{ij}\!\le\!0$. We replace $a_j$ with the same-strength attack action in both conditions and compare outcomes over an identical window $\mathcal W$. Per environment this yields $500$ base episodes $\times$ ordered pairs $\times 2$ conditions: {6{,}000} variants for $n{=}3$ (6 pairs), and {20{,}000} for $n{=}5$ (20 pairs). For the SMAC 3s\_vs\_3z, we experimented with $100$ base episodes, yielding {1{,}200} variants in total.

\vspace{.2cm}
\noindent\textit{Validation rationale (paired causal test).}
If $G_{ij}$ truly captures agent $j$'s capacity to affect agent $i$, then interventions at Critical moments (high $G_{ij}$ with $D_{ij}>0$) should induce more severe downstream instability than identical interventions at Robust moments (low $G_{ij}$ with $D_{ij}\leq 0$). We employ a paired experimental design where both intervention types are applied to the same agent pair $(j\to i)$ within the same episode. 

\vspace{.2cm}


\noindent\textit{Evaluation Metrics.}
To validate (\textsc{Patient-0} identification), we report four accuracy measures: Stage-1 accuracy $\Pr(\hat p_{\mathrm{S1}}{=}p^\star)$, Traceback accuracy $\%(\text{chain ends at }p^\star)$, Combined accuracy after Stage-2 correction, and Correction rate (the fraction of downstream-first misidentifications successfully corrected by Stage-2, calculated over all downstream-first cases). We didn't report \textsc{Patient-0} baselines because prior works address different problems: \cite{Lischke} detects per-timestep anomalies using LSTMs on state-action traces without identifying the initial failure source, while \cite{ijcai2023p19} develops decentralized GRU-based normality scoring for anomaly detection but does not identify the first inciting agent.

\begin{figure*}[htbp]
  \centering
  \begin{minipage}{0.94\textwidth}
  \centering

  \begin{subfigure}{0.77\linewidth}
    \centering
    \includegraphics[width=\linewidth]{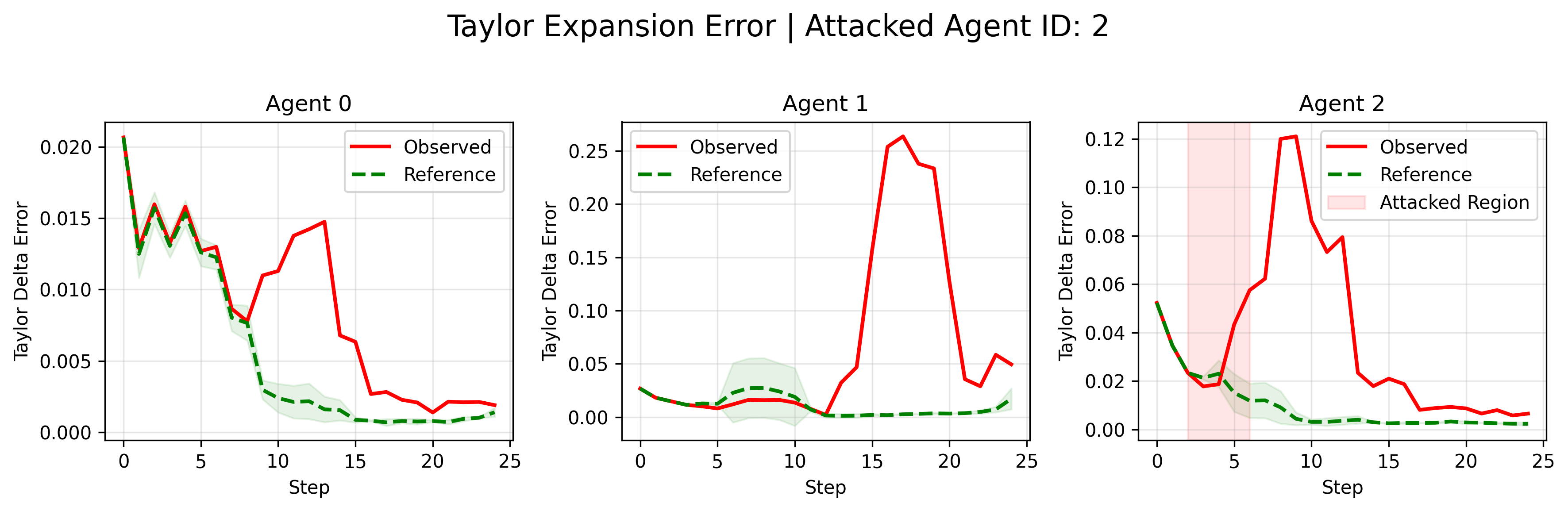}
    \caption{Stage~1: Taylor approximation error in all agents}
    \label{fig:high-taylor}
  \end{subfigure}

  \medskip

  \begin{subfigure}{0.45
  \linewidth}
    \centering
    \includegraphics[width=\linewidth]{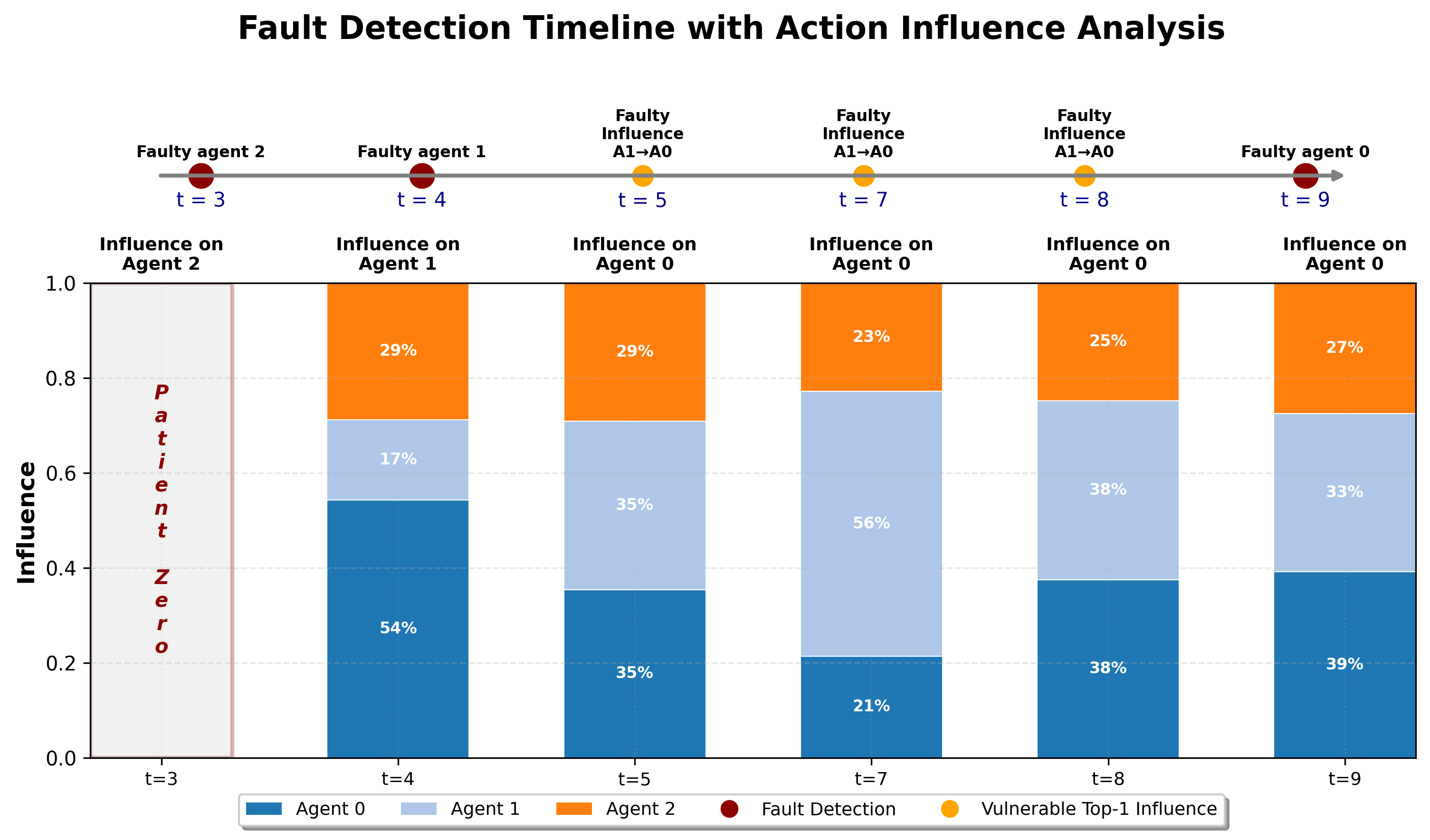}
    \caption{Stage~1,2: Influence timeline from the detection time of \textsc{Patient-0} to the detection of the last faulty agent}
    \label{fig:high-timeline}
  \end{subfigure}\hfill
  \begin{subfigure}{0.45\linewidth}
    \centering
    \includegraphics[width=\linewidth]{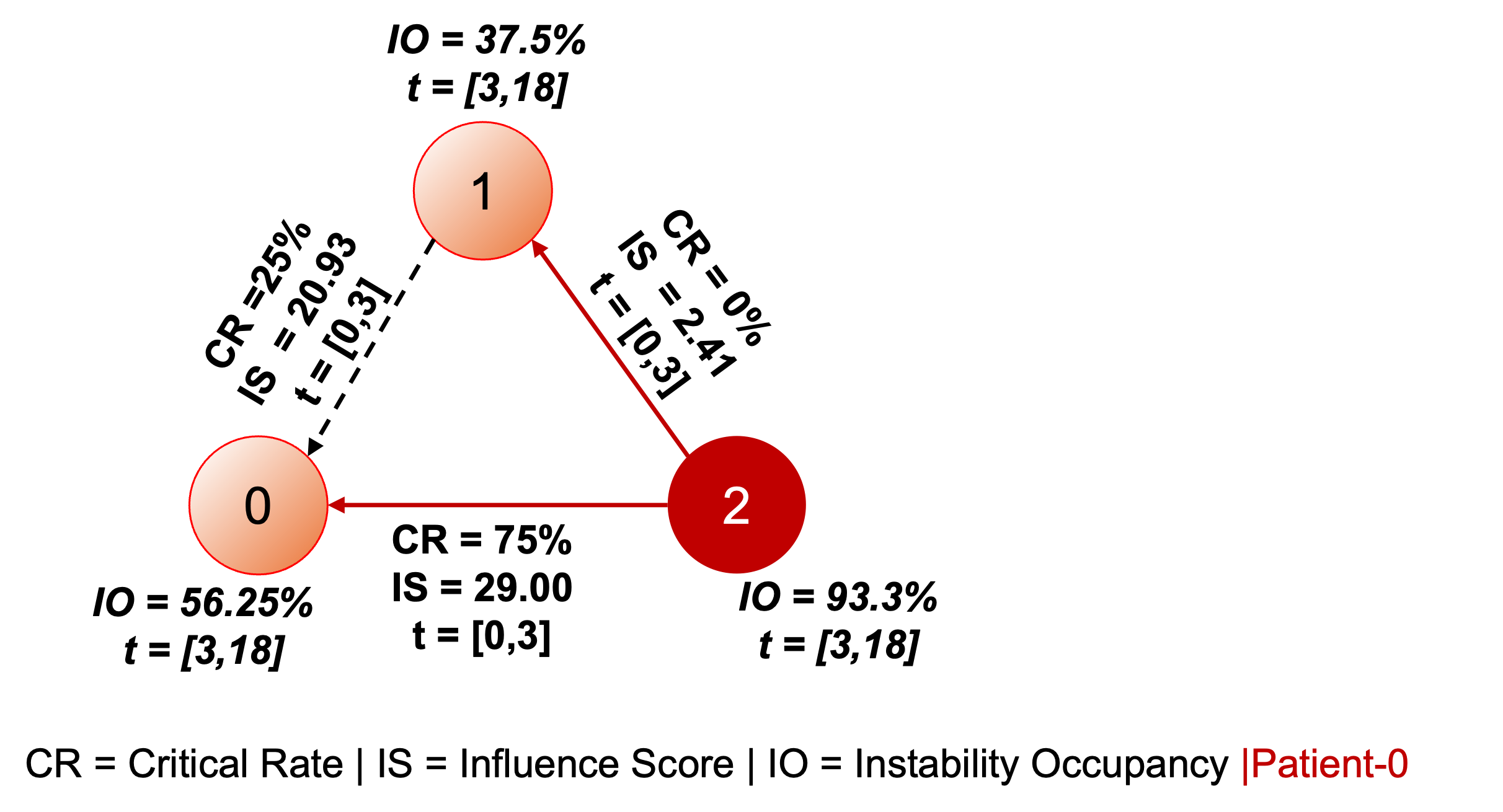}
    \caption{Stage~1\&2: Contagion graph. Nodes with gradient color represent faulty agents}
    \label{fig:high-graph}
  \end{subfigure}

  \caption{\textbf{High-influence episode.} Top: Stage~1 Taylor-error signals (three agents). Bottom: stacked influence timeline (left) and Stage~2 contagion graph (right), showing dominant accelerating pathways.
  \emph{Notation for time windows:} The label \(t[\cdot,\cdot]\) on \textbf{nodes} denotes the interval used to evaluate the node’s Instability Occupancy (IO): for agent \(i\) with detection time \(T_i\), we observe IO from \(t[T_i,\;T_i{+}15]\). On \textbf{edges} \(j\!\to\! i\), the “recent~5” tag denotes the last up-to-5 steps ending at the downstream detection time \(T_i\), i.e., \(\{ \max(0,\,T_i{-}4),\ldots,T_i\}\), which are used to compute the edge’s Critical Rate (CR) and Influence Score (IS).}
  \label{fig:high-influence-rows}
\end{minipage}
\end{figure*}

\begin{figure*}[htbp]
  \centering
  \begin{minipage}{0.95\textwidth}
  \centering
  \begin{subfigure}{0.8\linewidth}
    \centering
    \includegraphics[width=\linewidth]{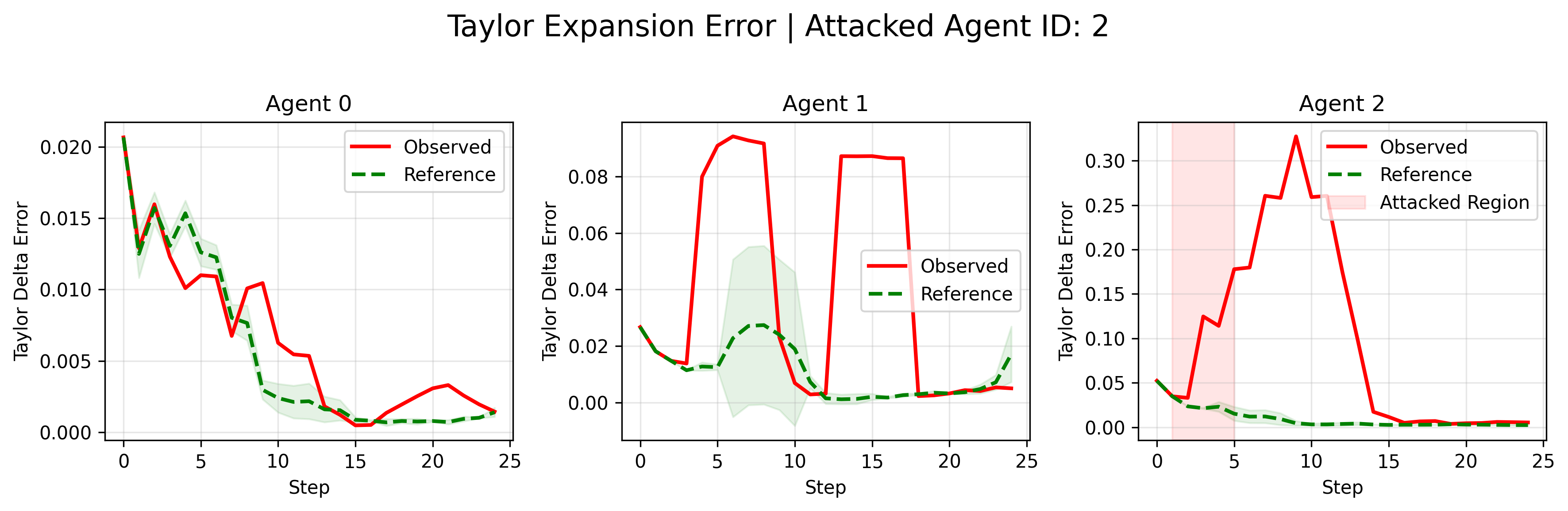}
    \caption{Stage~1: Taylor approximation error in all agents.}
    \label{fig:low-taylor}
  \end{subfigure}

  \medskip

  \begin{subfigure}{0.45\linewidth}
    \centering
    \includegraphics[width=\linewidth]{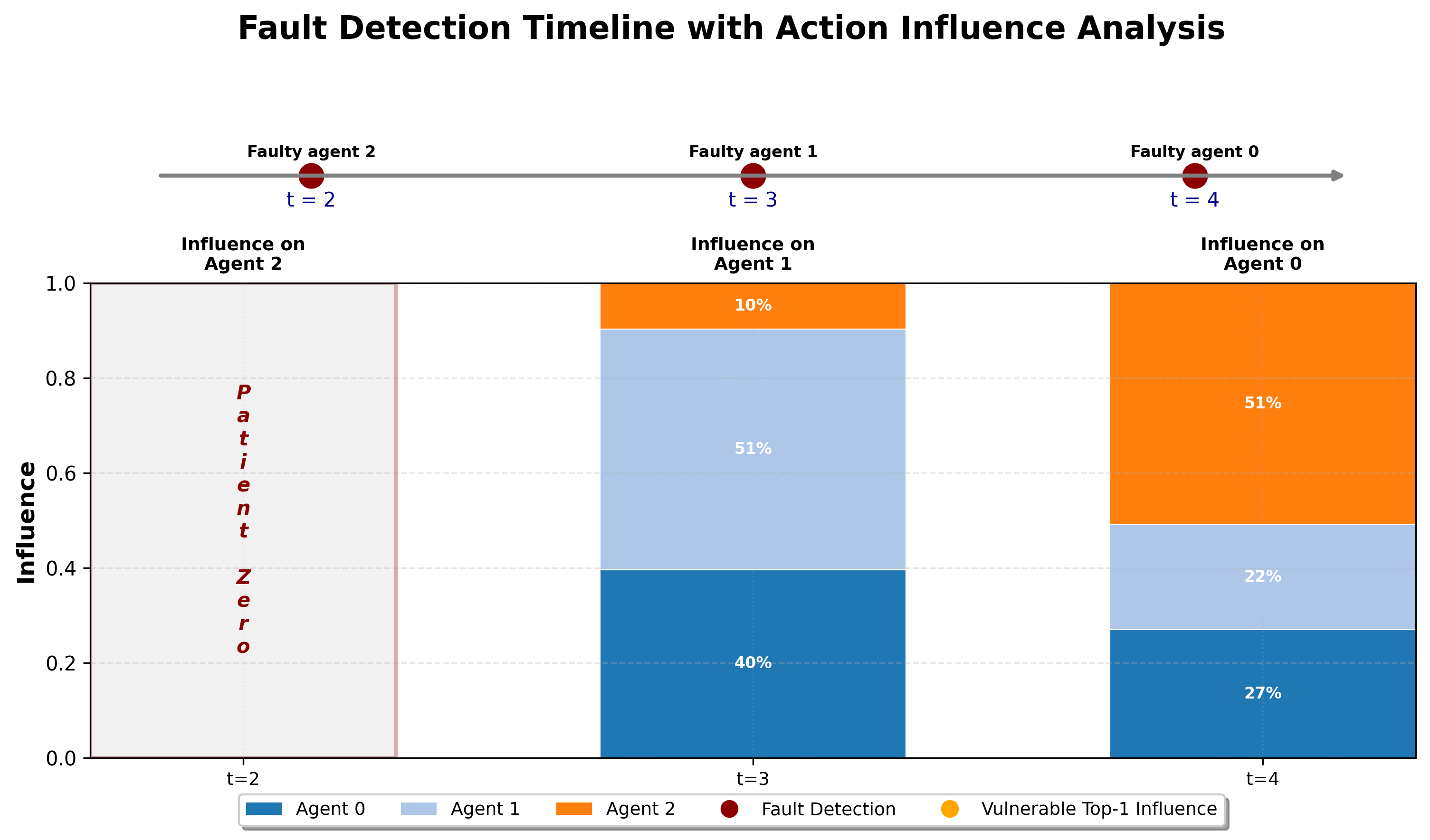}
    \caption{Stage~1\&2: Influence timeline from the detection of \textsc{Patient-0} to the detection of the last faulty agent.}
    \label{fig:low-timeline}
  \end{subfigure}\hfill
  \begin{subfigure}{0.45\linewidth}
    \centering
    \includegraphics[width=\linewidth]{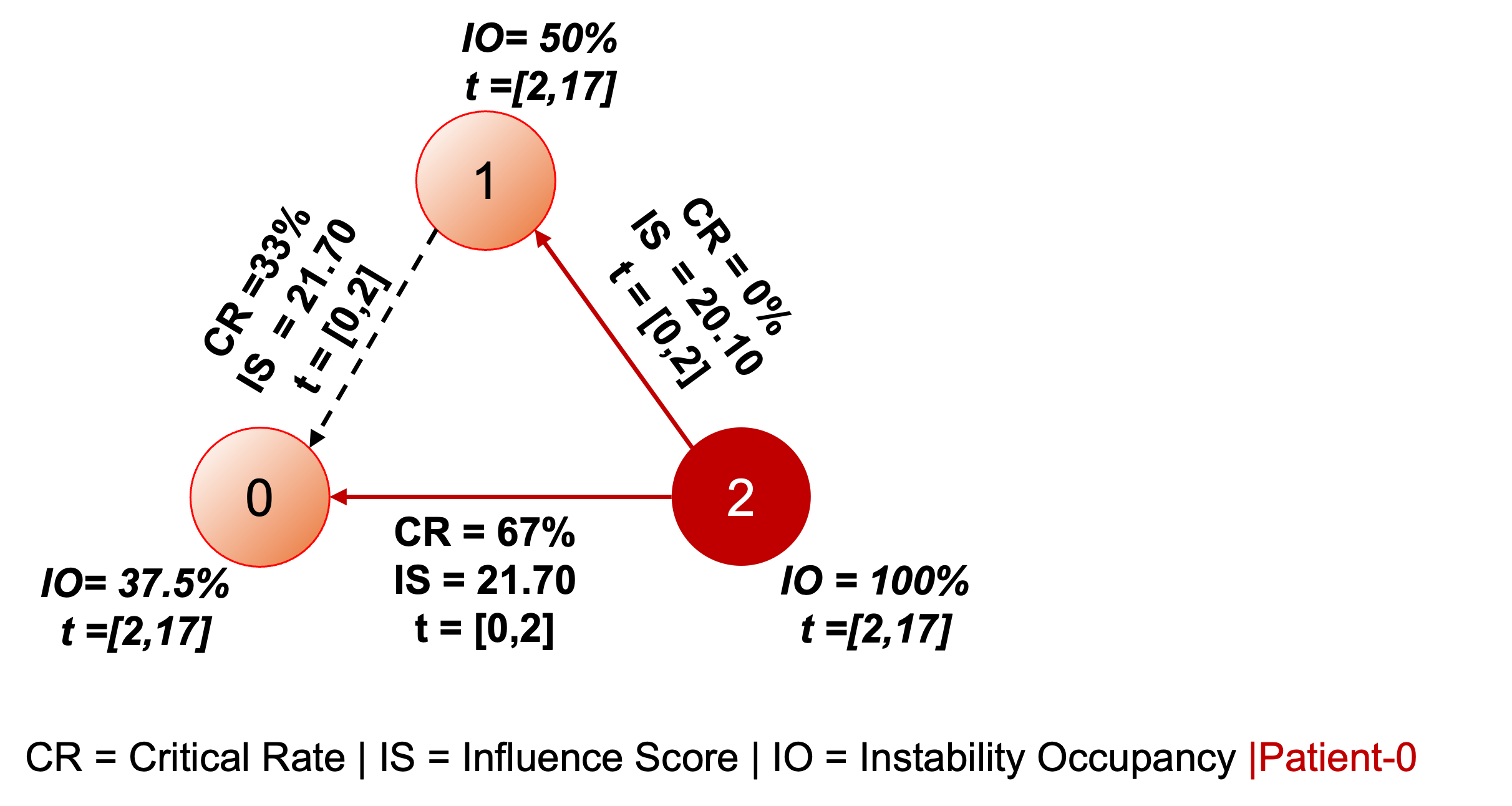}
    \caption{Stage~1\&2: Contagion graph. Nodes with gradient color represent faulty agents.}
    \label{fig:low-graph}
  \end{subfigure}

  \caption{\textbf{Low-influence episode.} Top: Stage~1 Taylor-error signals (three agents). Bottom: influence timeline (left) and Stage~2 contagion graph (right), illustrating weaker upstream pathways.}
  \label{fig:low-influence-rows}
  \end{minipage}
\end{figure*}

\subsection{Results}
\noindent \textbf{Results for \textsc{Patient-0} iIdentification (Table.~\ref{tab:p0-results})}. Stage-1 accuracy shows HATRPO consistently outperforming MADDPG ($\Delta$+3.4\% to $\Delta$+10.8\% across settings), confirming our hypothesis that smoother gradient landscapes (HATRPO) yield cleaner Taylor-error signals. MADDPG's exploration noise creates higher gradient jitter, particularly damaging in larger teams (Simple Spread-5: 88.1\% vs. HATRPO's 98.9\%).

Stage-2 correction rates reveal the method's strength in coordinated settings: SMAC achieves 70.8-76.9\% correction by leveraging tight coupling dynamics, while Simple Spread-5 MADDPG struggles (40.1\%) with diffuse influence across 5 agents. This pattern precisely matches our theoretical expectation: directional curvature effectively traces amplification pathways but requires sufficiently strong coupling signals. The stark contrast between environments highlights this dependency: Simple Spread's cooperative navigation creates relatively weak, distributed couplings compared to SMAC's real-time strategy combat, where targeting decisions, focus fire, and ability coordination create intense, immediate interdependencies that our directional curvature term readily detects.

Combined accuracy demonstrates the two-stage design working as intended: substantial gains where Stage-1 struggles (MADDPG: +4.7-8.2\% improvements) while maintaining near-perfect performance where Stage-1 already excels.

\noindent \textbf{Results for Influence Validation (Tabel.~\ref{tab:influence-validation})}.  Our instability occupancy (IO) metric consistently outperforms traditional performance measures (drop in reward and Q) by 20+ percentage points in most cases (e.g., Simple Spread-3 MADDPG: 77.7\% vs. 51.2\%/48.4\%). Critically, IO remains decisively above chance even when AUC metrics fail, proving gradient geometry captures vulnerability signals invisible to performance-based approaches.

Algorithmic patterns reveal how underlying RL methods affect our metric's fidelity: HATRPO maintains strong IO accuracy even in complex settings (Simple Spread-5: 82.4\%, SMAC: 59.7\%) while MADDPG degrades with task complexity (SMAC: 54.5\%). This directly aligns with our method's dependence on stable critic landscapes and precise action gradients. HATRPO's trust regions and advantage normalization provide these, while MADDPG's exploration noise and continuous relaxations of discrete actions introduce noise that corrupts our gradient-based signals.

The MADDPG results reveal a key strength of our two-stage design: even with modest Stage-1 IO accuracy (54.5\%), Stage-2 achieves strong correction rates (70.8\%). This demonstrates that curvature gating can salvage imperfect detection. While Stage-1 may misidentify exactly which agents are unstable, Stage-2's directional curvature analysis still correctly traces the true amplification pathways through SMAC's strong, structured couplings. The method proves robust to noisy inputs when the underlying influence topology is clear.

\noindent\textbf{Remarks.}
(i) When gradient landscapes are smooth (HATRPO), Stage--1 alone is highly accurate; Stage--2 primarily fixes the occasional downstream-first episode (large boost on SMAC). (ii) When gradients are jittery or influence is diffuse (MADDPG in Simple Spread-5), extend the traceback window and allow multi-parent edges to capture split pathways. (iii) For discrete-action tasks like SMAC, using temperature-scheduled Gumbel--Softmax (or an auxiliary probe critic) can improve derivative fidelity and further strengthen IO and correction rates.
\graphicspath{{visualization/}}

\begin{figure*}[htbp]
  \centering

  \begin{minipage}{0.94\textwidth}
  \centering

  \begin{subfigure}{0.70\linewidth}
    \centering
    \includegraphics[width=\linewidth]{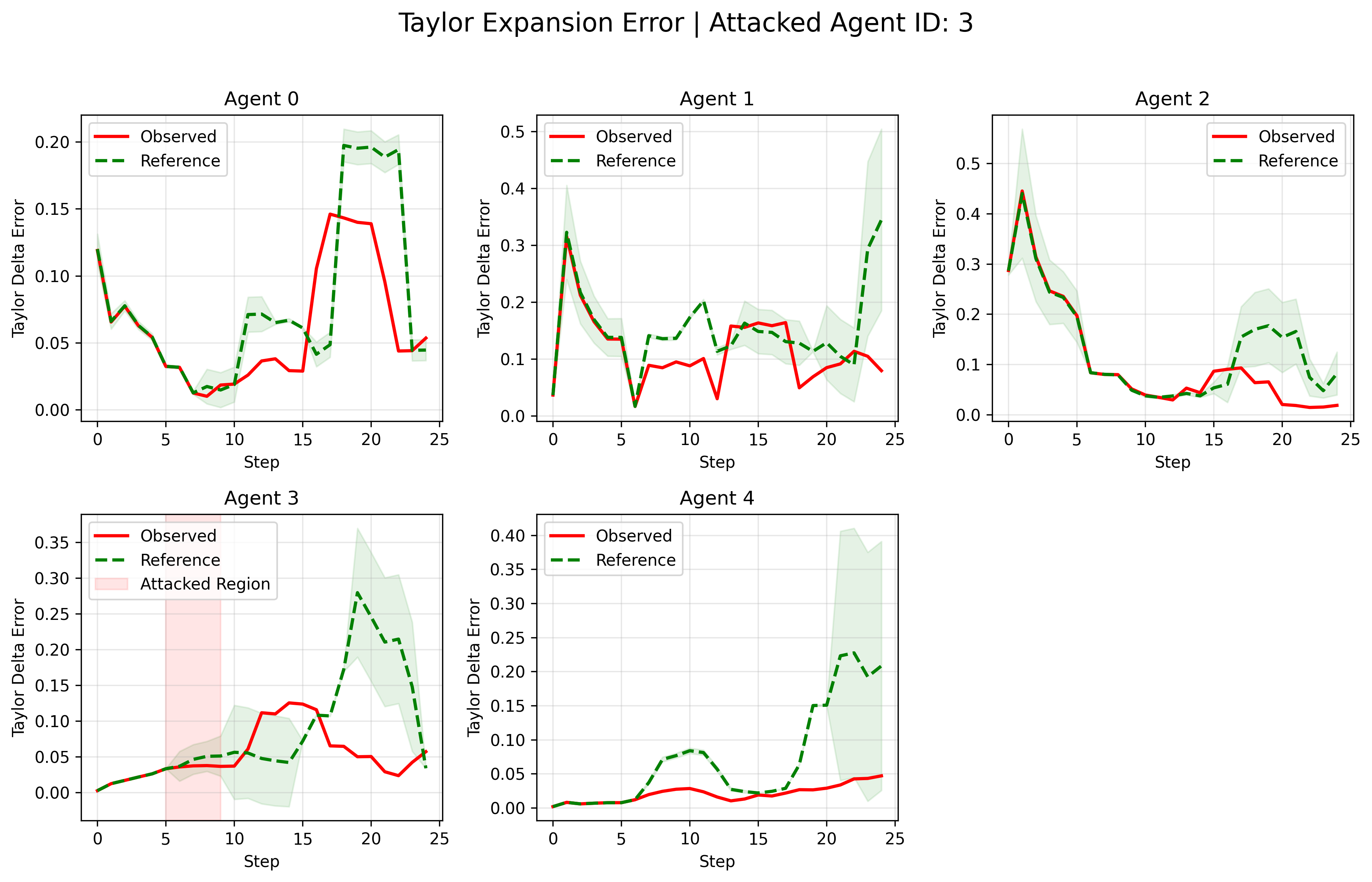}
    \caption{Stage~1: Taylor-approximation error across agents (downstream-first episode).}
    \label{fig:falsep0-taylor}
  \end{subfigure}

  \medskip

  \begin{subfigure}{0.45\linewidth}
    \centering
    \includegraphics[width=\linewidth]{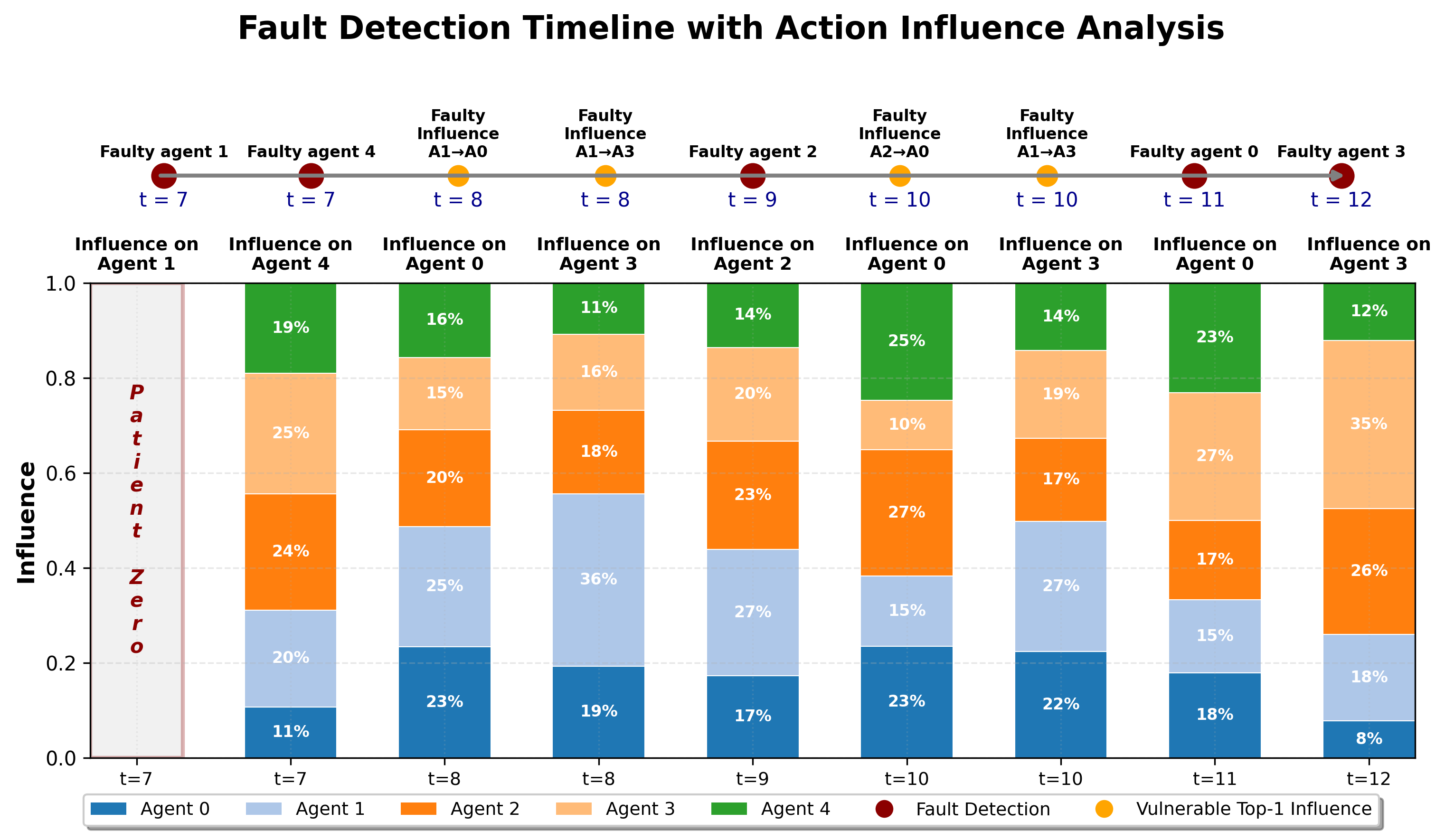}
    \caption{Influence timeline around detections (stacked upstream shares).}
    \label{fig:falsep0-timeline}
  \end{subfigure}\hfill
  \begin{subfigure}{0.45\linewidth}
    \centering
    \includegraphics[width=\linewidth]{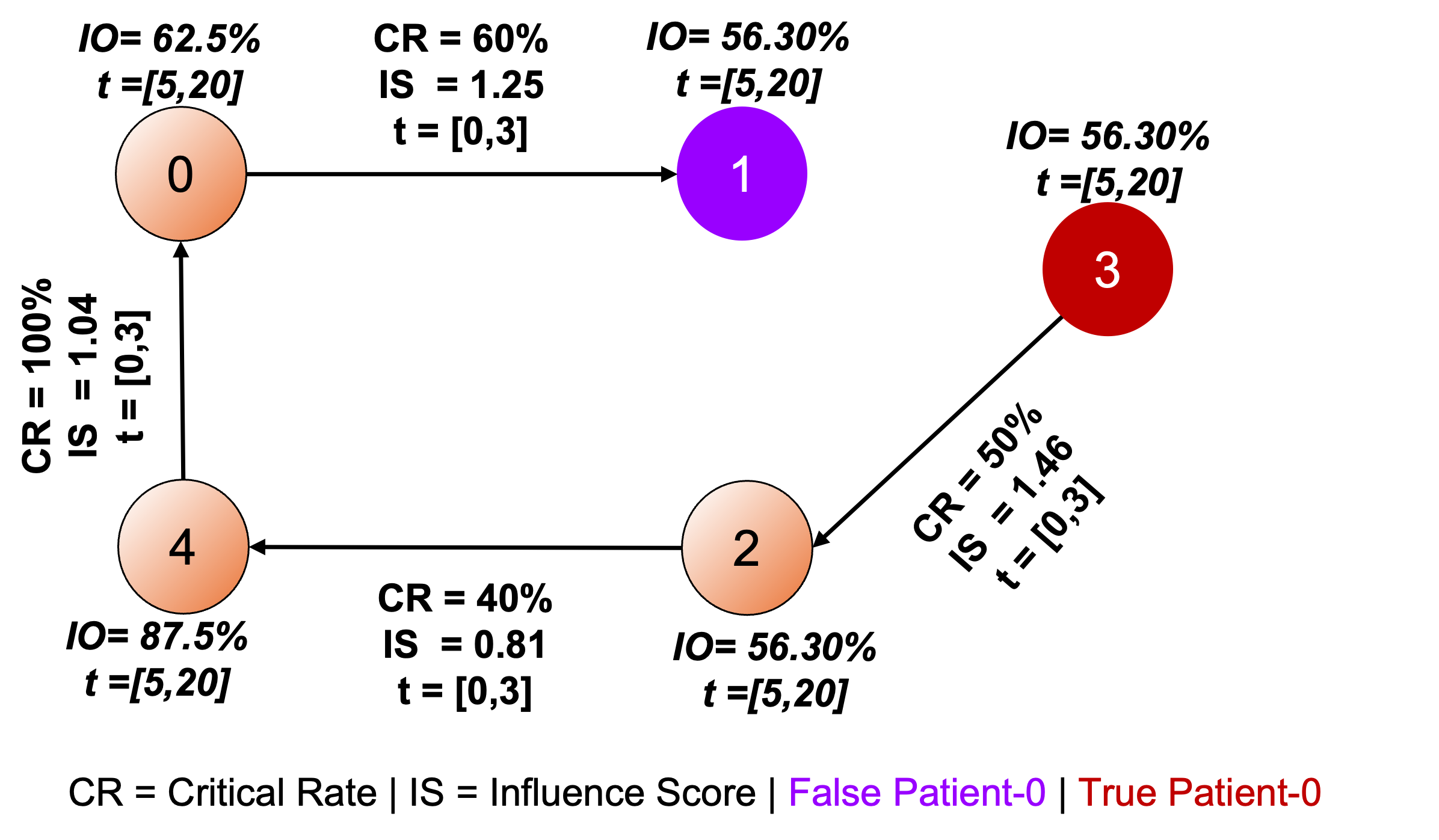}
    \caption{Stage~2: traceback/contagion graph correcting the false \textsc{Patient-0}. Nodes with gradient fill are faulty.}
    \label{fig:falsep0-graph}
  \end{subfigure}

  \caption{\textbf{False \textsc{Patient-0} (downstream-first) visualization.}
  \emph{Top:} Stage~1 Taylor-error signals lead to an early flag on a downstream agent.
  \emph{Bottom-left:} Influence timeline shows strong upstream contributions near each detection.
  \emph{Bottom-right:} Stage~2 traceback via directional critic curvature reveals the true source and the amplifying pathway.
}
  \label{fig:falsep0-rows}

  \end{minipage}
\end{figure*}

\subsection{Contagion Graph for Interpretable Failure Analysis.}
We present two representative cases that show how the two-stage method detects, explains, and corrects failure dynamics. Throughout, Instability Occupancy (IO) is the ``Exceed Rate'' in the figures, and edge annotations use Critical Rate (CR) for the frequency of accelerating states ($D_{ij}\!>\!0$) and Influence Score (IS) for cumulative first-order sensitivity (our aggregated $G$ signal). The visuals are organized to align with our pipeline: Stage~1 evidence (Taylor-error and IO timelines) on the left, and Stage~2 evidence (accelerating edges and traceback) on the right.

\vspace{.2cm}

\noindent\textbf{Case 1 (Simple Spread, $n{=}3$, MADDPG, edge $2{\to}0$).}
We perform two one-shot interventions on the same edge $2{\to}0$ with the \emph{same} attack strength and post-intervention horizon $K$: 
(i) a \emph{Critical} moment where the local sensitivity is high (large $G_{2\to0}$) and the directional curvature is \emph{accelerating} ($D_{2\to0}{>}0$), and 
(ii) a \emph{Robust} moment where sensitivity is low and/or curvature is non-accelerating ($D_{2\to0}{\le}0$). 
Figures~\ref{fig:high-influence-rows} (high influence) and \ref{fig:low-influence-rows} (low influence) visualize the resulting Stage~1 and Stage~2 signals.

\noindent \underline{Stage~1: What becomes unstable (and for how long).} 
In the high-
influence episode (Fig.~\ref{fig:high-influence-rows}a), the downstream agent~0 exhibits a clear Taylor-error surge relative to its reference trajectory, and the excursion persists over many steps. 
This is reflected in a larger instability occupancy (IO) for node~0 over the node window $t[T_0,\,T_0{+}15]$ (shown on the node label in the graph). 
In contrast, in the \emph{low-influence} episode (Fig.~\ref{fig:low-influence-rows}a), the Taylor-error deviations at agent~0 are notably smaller and shorter, yielding a lower IO over the same evaluation window.
Thus, Stage~1 indicates that striking $2{\to}0$ at a high-$G$ accelerating state produces stronger and more persistent local instability than striking it at a robust moment.

\noindent \underline{Stage~2: Why it happens (and from where).}
The stacked influence timelines (Figs.~\ref{fig:high-influence-rows}b and \ref{fig:low-influence-rows}b) align contributions around detection and show how much of agent~0's update pressure is attributable to each teammate. 
In the high-influence case, agent~2 contributes a dominant upstream share to agent~0 near $T_0$, consistent with a propagating cascade. 
The contagion graphs (Figs.~\ref{fig:high-influence-rows}c and \ref{fig:low-influence-rows}c) make this causal picture explicit using our directional metrics computed on the \emph{recent-5} edge window $\{\max(0,T_0{-}4),\dots,T_0\}$ (shown on each edge label):
the edge $2{\to}0$ carries a higher Critical Rate (CR; fraction of steps with $D_{2\to0}{>}0$) and a larger Influence Score (IS; masked accumulation of $G_{2\to0}$ over steps with $D{>}0$) in the high-influence episode, but both CR and IS are lower in the low-influence episode. 
Node labels simultaneously display the IO window $t[T_i,\,T_i{+}15]$, tying the downstream instability back to its temporal evaluation range.

\noindent Viewed together, the episodes reveal the mechanism: perturbing the source at a high-$G$, $D{>}0$ state produces stronger, longer downstream instability that Stage~2 traces to an accelerating path (high CR/IS on $2{\to}0$); perturbing the same edge at a robust (low-$G$ or $D{\le}0$) state yields muted, non-propagating effects (low CR/IS).

\noindent\textbf{Case 2 - Downstream-first flag with traceback correction (Simple Spread, $n{=}5$, MADDPG).}
An adversarial perturbation is applied to agent~3 over $t{=}5\text{--}8$. Yet Stage~1 raises the first alarm at agent~1 at $t{=}7$, creating a downstream-first cascade and a false \textsc{Patient-0}.

\noindent\underline{Stage~1: What trips first (and where instability persists).}
Figure~\ref{fig:falsep0-taylor} shows agent~1’s Taylor-error crossing the threshold earliest at $t{=}7$, while agents~0 and~2 exhibit sub-threshold rises and agent~3’s local error remains comparatively muted during the shaded attack window. The instability-occupancy (IO) windows $t[T_i,\,T_i{+}15]$ (shown on node labels in Fig.~\ref{fig:falsep0-graph}) confirm that agent~1 accrues the earliest and most sustained excursion despite being downstream.

\noindent\underline{Stage~2: Why agent~1 fires first (and where amplification began).}

\noindent Aligned to $T_1$, the stacked influence timeline in Fig.~\ref{fig:falsep0-timeline} concentrates upstream pressure from $\{3,2,0\}$ onto agent~1 immediately before detection. The contagion graph in Fig.~\ref{fig:falsep0-graph} gates edges by accelerating directional curvature ($D_{ij}{>}0$ over the \emph{recent-5} steps before $T_i$) and accumulates sensitivity-weighted mass as an Influence Score (IS). This traceback identifies agent~3 as the \emph{true} \textsc{Patient-0} and explains why the receiver (agent~1) crossed the threshold first.


\noindent\underline{Takeaway.}
Stage~1 alone may nominate a false \textsc{Patient-0} when amplification concentrates on a downstream receiver. Stage~2 corrects this failure mode by (i) gating for accelerating edges ($D{>}0$), (ii) weighting by sensitivity (IS), and (iii) aggregating over the \emph{recent-5} causal window recovering the true source (agent~3) and the pathway that carried the pressure to agent~1.

\section{Discussion and Conclusion}

We introduce, to our knowledge, the first framework for \textsc{Patient-0} detection and influence traceback in MARL. By turning opaque multi-agent failures into critic–geometric evidence and contagion graphs, the method yields actionable forensics for safety-critical deployments: it identifies root causes, explains detection anomalies, and maps how failures propagate. Our approach is a compact two-stage procedure: Stage~1 flags per-agent onsets from Taylor-error anomalies and summarizes persistence (IO); Stage~2 traces causal routes via directional critic derivatives, emphasizing states with high sensitivity and accelerating curvature.
Limitations include a single-source of failure assumption; reliance on differentiable critics and approximate gradient–influence alignment; sensitivity to thresholds and windowing; and added computational cost (gradients and Hessian–vector products). Future work will relax the single-source assumption, reduce overhead with lighter estimators, harden robustness to critic misspecification, and develop online detection at scale on larger, more heterogeneous teams and real deployments.

\begin{acks}
This material is based upon work supported by the National Science Foundation (NSF) under grant no. $2442581$.
\end{acks}

\bibliographystyle{ACM-Reference-Format} 
\balance
\bibliography{sample}


\end{document}